\documentclass[letterpaper, 10 pt, conference]{ieeeconf}  % Comment this line out if you need a4paper

\IEEEoverridecommandlockouts                              % This command is only needed if 
\usepackage{graphics} % for pdf, bitmapped graphics files
\usepackage{epsfig} % for postscript graphics files
\usepackage{mathptmx} % assumes new font selection scheme installed
\usepackage{times} % assumes new font selection scheme installed
\usepackage{amsmath} % assumes amsmath package installed
\usepackage{amssymb}  % assumes amsmath package installed
\usepackage{hyperref}
\usepackage{booktabs}
\usepackage{mathtools}
\usepackage{graphicx,tabularx}
\usepackage{svg}
\usepackage{multirow}
\usepackage{blindtext}  

\usepackage{subcaption}

\usepackage{xcolor}

\title{\LARGE \bf
A Methodology for Designing Knowledge-Driven Missions for Robots
}
\author{Guillermo GP-Lenza, Carmen DR.Pita-Romero, Miguel Fernandez-Cortizas, Pascual Campoy \\
Computer Vision and Aerial Robotics group (CVAR), Centre for Automation and Robotics (CAR) \\ Universidad Politécnica de Madrid (UPM),  Madrid, Spain. % <-this % stops a space
\thanks{Corresponding author: g.glenza@upm.es }
}
\begin{document}

\maketitle
\thispagestyle{empty}
\pagestyle{empty}

%%%%%%%%%%%%%%%%%%%%%%%%%%%%%%%%%%%%%%%%%%%%%%%%%%%%%%%%%%%%%%%%%%%%%%%%%%%%%%%%
\color{black}
\begin{abstract}
This paper presents a comprehensive methodology for implementing knowledge graphs in ROS 2 systems, aiming to enhance the efficiency and intelligence of autonomous robotic missions. The methodology encompasses several key steps: defining initial and target conditions, structuring tasks and subtasks, planning their sequence, representing task-related data in a knowledge graph, and designing the mission using a high-level language. Each step builds on the previous one to ensure a cohesive process from initial setup to final execution. A practical implementation within the Aerostack2 framework is demonstrated through a simulated search and rescue mission in a Gazebo environment, where drones autonomously locate a target. This implementation highlights the effectiveness of the methodology in improving decision-making and mission performance by leveraging knowledge graphs.
% \blindtext % REMOVE

\end{abstract}
%%%%%%%%%%%%%%%%%%%%%%%%%%%%%%%%%%%%%%%%%%%%%%%%%%%%%%%%%%%%%%%%%%%%%%%%%%%%%%%%
\section*{SUPLEMENTARY MATERIAL}
Video of the experiments: \url{https://vimeo.com/cvarupm/knowledgeformissions} \\
Released code: \url{https://github.com/aerostack2/as2_knowledge_graph}

\color{black}
\section{INTRODUCTION}

The rapid advancement of Artificial Intelligence (AI) technology has significantly expanded the applications of robotics, particularly in the field of mobile robotics. These mobile robotic systems are increasingly utilized in diverse domains such as agriculture, logistics, surveillance, environmental monitoring, and search and rescue operations. As these robots operate in complex, dynamic, and often unpredictable environments, there is a growing need to enhance their autonomy to perform tasks with higher accuracy, efficiency, and adaptability.

One of the critical factors in achieving greater autonomy in mobile robotic systems is the effective use of knowledge. Mobile robots require comprehensive and precise knowledge about their environment, tasks, actions, and inherent capabilities to make informed decisions and execute tasks successfully across different contexts. Traditionally, this knowledge has been encoded in algorithms and static databases, which often lack the flexibility and scalability needed to handle the dynamic nature of real-world scenarios.

To address these limitations, the concept of Knowledge Graphs (KGs) has emerged as a powerful tool for knowledge representation and reasoning. KGs provide a structured and semantically rich framework for organizing information, enabling both the representation of complex relationships and efficient algorithms for knowledge retrieval and reasoning. Integrating KGs into mobile robot systems can enhance their capability to perform these tasks autonomously by providing real-time situational awareness, contextual understanding, and decision support.

Despite these advantages, the development and implementation of KGs in mobile robotic systems present several challenges, including data integration from various sensors and sources, real-time knowledge updating, or multi-agent interaction.

This paper aims to develop a comprehensive methodology for applying KG to existing ROS 2 based robotic systems using a "brownfield" approach. Our goal is to enhance the explainability of the robot's operations during various stages and to leverage this knowledge for informed decision-making processes.

To this end, we provide the following contributions:
\begin{itemize}
    \item A detailed description of each step of the methodology, specifying the required inputs and the process to achieve the desired outputs to ensure clarity and precision in executing each phase effectively.

    \item Software tools to apply the described methodology to any ROS 2 based robotic system.

    \item A working example describing the application of the proposed methodology to Aerostack2 \cite{aerostack2}, a ROS 2 open-source framework to design and control aerial robotic systems.
\end{itemize}

\subsection{Related work}

Mobile robotic systems have seen significant advancements with the integration of knowledge-based systems, which enhance their decision-making and adaptability in dynamic environments. Various approaches have been explored in this domain, each one leveraging distinct methodologies and technologies to improve robotic autonomy and efficiency. This section delves into the diverse landscape of mobile robotic systems incorporating knowledge-based systems, highlighting key innovations, methodologies, and their respective contributions to the field.

Cognitive architectures are comprehensive, computational frameworks designed to model the structures and processes of human cognition. These architectures serve as blueprints for understanding and replicating the intricacies of human thought, encompassing perception, memory, reasoning, and learning. They provide a unified platform for developing intelligent agents capable of performing complex tasks, offering significant advancements in fields ranging from robotics to human-computer interaction.

\textbf{SOAR} \cite{SOARPaper}, was the first cognitive architecture integrated into real robots and used with multiple robots \cite{benjamin2006embodying}. Its basic architecture is established by representing the state of an environment through a graph composed of discrete objects and continuous properties. This allows for a set of predicates to be independent and fixed within the architecture, while the decisions regarding which predicates should be extracted are determined by the specific task that an agent must perform \cite{laird2012cognitive}.

\textbf{ACT-R} \cite{ACTRPaper} models human cognition by integrating symbolic and subsymbolic processing. Symbolic processing includes declarative memory, which stores knowledge as chunks (data and facts) with labels (slots), and procedural memory, which holds production rules in "if-then" statements to guide behavior based on current goals. Subsymbolic processing operates using a connectionist model, resolving conflicts by selecting the chunk with the highest activation level, determined by past utility and context relevance. Through cycles of perception, cognition, and action, ACT-R adapts to changing environments and tasks, effectively storing and managing knowledge.

\textbf{LIDA} \cite{LIDAPaper}, provides adaptation and continuous learning by utilizing a multilayer working memory system. Each layer within this system has a distinct purpose and stores various types of information: perceptual, declarative, memory, and procedural. Active working memory serves as the interface that connects all these layers, allowing for the manipulation of stored information. LIDA employs a distributed representation where information is encoded through the activation patterns of artificial neural networks, offering a robust mechanism for adapting to dynamic environments. 

Aside from cognitive architectures, other systems focus more specifically on concrete knowledge representation methods. These methods, used in modern advancements in artificial intelligence, machine learning, and neural networks, often emphasize structured data storage, pattern recognition, and internal learned representations. 

The proposed ontology structure by \cite{Sinpar} comprises three hierarchical layers where each layer regards more specific knowledge than the former: a metaontology that represents generic concepts, an ontology schema defining domain-specific knowledge, and an ontology instance that captures specific information about individual objects and their attributes. These layers are organized into six classes: Feature, Object, Actor, Space, Context, and Action, each with varying levels of detail. This structure provides a comprehensive, object-oriented, and frame-based language while the hierarchical structure that allows for knowledge to be effectively used through reasoning. 

%Bidirectional reasoning allows mutual information transfer between classes, properties, and instances for consistent knowledge representation whilst unidirectional reasoning across different layers integrates diverse information, enhancing the system's ability to apply this knowledge in various scenarios and tasks. 

The system proposed in \cite{LearningKBS} utilizes a knowledge-based system that integrates explicit expert knowledge with implicit learned knowledge, allowing the system to update its model based on the acquired information continuously. Through a knowledge acquisition module explicit knowledge provided by the operator is captured and converted into machine-readable form and then integrated with implicit knowledge. Implicit knowledge is captured by training a model to imitate the adjustments made by the operator, ultimately leading to full automation of robot programming. By learning from the operator's adjustments, the system enhances its flexibility, adaptability, and performance, addressing the challenges of industrial robot programming and improving overall efficiency in production processes.

Knowledge Graphs (KG) were first introduced in \cite{FirstKG} and later popularized by Google, and have evolved significantly, becoming powerful tools for structuring and managing complex relationships between data in various fields, including autonomous systems and robotics. In \cite{HierarchicalEmbeddKG}, authors discuss the use of KGs in enhancing robot manipulation tasks. They introduce a multi-layer knowledge-representation model that incorporates various elements such as scenes, objects, agents, tasks, actions, and skills. This hierarchical structure allows for a more nuanced understanding of manipulation tasks compared to traditional flat representations. The authors propose a heterogeneous graph-embedding method that assigns different weights to various relations within the KG to enhance reasoning capabilities. This approach allows the system to differentiate the significance of different connections, facilitating more nuanced reasoning about how various factors influence manipulation tasks.

Compared to traditional cognitive architectures, KGs offer a more flexible and scalable approach to representing information. While cognitive architectures are highly specialized in replicating human-like reasoning and learning, they are limited in adaptability across various tasks or domains. KGs, in contrast, provide a dynamic, interconnected representation of entities and their relationships, allowing for more granular, real-time information querying and updating.

\color{black}
% \newpage
% \newpage

% \section{DYNAMIC TRAJECTORY GENERATION}

\section{METHODOLOGY}

\begin{figure*}[htb]
    \centering
    \includegraphics[width=.95\linewidth]{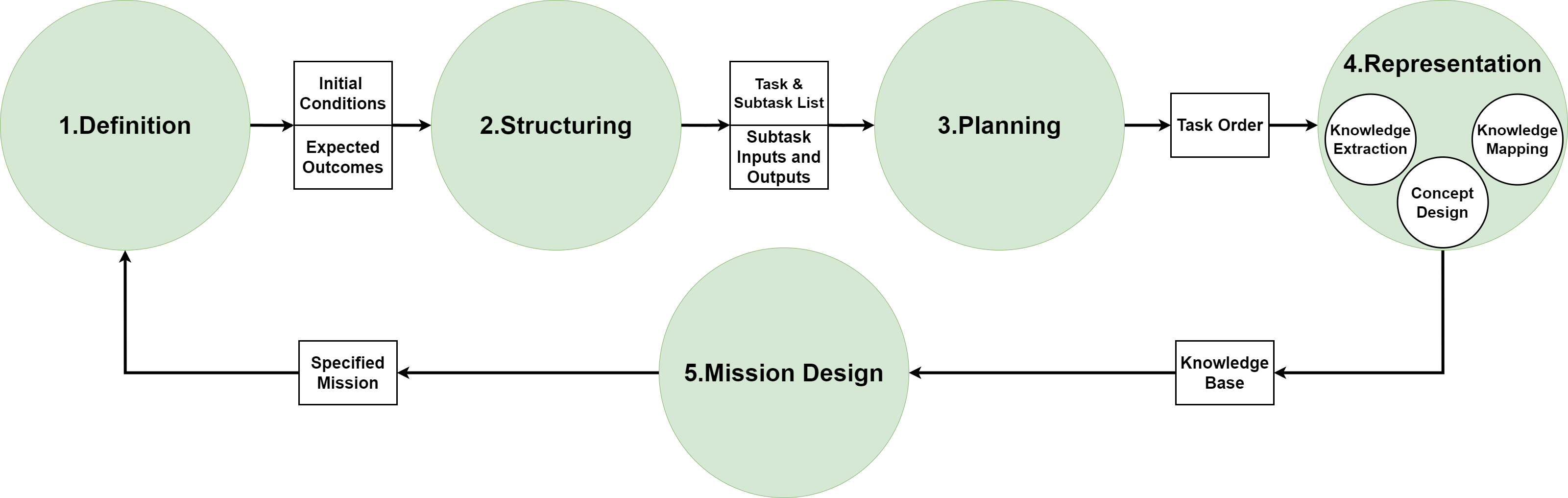}
    \caption{A full overview of the described methodology. Circles represent each step described in the proposed methodology while rectangular boxes contain the outcomes of each step.}
    \label{fig:graphmethodo}
\end{figure*}

The proposed methodology's objective is to integrate KGs into ROS 2 systems providing a structured approach for leveraging the full potential of KGs, thus enabling more informed decision-making and improved mission performance in diverse ROS 2 based applications. This methodology is composed of several key steps: defining initial and target conditions, structuring tasks and sub-tasks, planning their sequence, representing task-related data in a KG, and designing the mission using a high-level language. Each step builds on the previous one, ensuring a cohesive process from initial setup to final execution. A full overview of the methodology is shown in Figure \ref{fig:graphmethodo}.

\subsection{Definition}
In the definition step, initial conditions of the mission and expected outcomes are defined. This foundational step involves a thorough understanding of the mission's goals and constraints, providing a clear vision of the desired outcomes. This step aims to establish a baseline against which the subsequent steps will be measured, ensuring that all efforts align with the ultimate mission objectives.

\subsection{Structuring}
The structuring step involves breaking down the mission into a detailed list of tasks necessary to achieve the defined target conditions. Each task is further subdivided into sub-tasks, with a focus on identifying and specifying the inputs and outputs associated with each sub-task, helping to understand essential activities, their interconnections, and decision points necessary for mission success. These inputs and outputs form the foundational elements necessary for a robust mission and will be mapped to the KG in a later step.

% In this stage, the mission is organized by sequentially listing and detailing the tasks required. Flowcharts or other visual tools are used to represent the sequence of tasks, 

\subsection{Planning}
Once the tasks and sub-tasks are defined, the planning step requires establishing a valid sub-tasks ordering. This ordering should reflect a logical sequence that ensures all prerequisites are met before moving on to subsequent tasks. The primary goal in this step is to verify that the proposed sequence will effectively lead to the achievement of the mission’s target conditions as outlined in the definition step. A well-ordered plan serves as a roadmap for the subsequent stages, facilitating smooth execution and integration.

\subsection{Representation}
The representation step involves mapping the inputs and outputs of each sub-task to a KG representation. This step is critical for translating the relevant data identified through the task list into a form that can be effectively utilized within the KG framework. By aligning the inputs and outputs with the KG, it is ensured that the necessary information is available and properly organized for efficient querying and control.

\subsubsection{Knowledge Extraction}
To apply this methodology to any given ROS 2 based system, it is essential to design a data extraction method tailored to the system’s specific architecture, data sources and domain ensuring accurate and efficient information retrieval.

\subsubsection{Concept Design}
During the concept design step, the designer should define how the data will be represented as distinct entities and identify the relationships that connect these entities in the KG. Accurately mapping the data to the KG is essential to ensure that the KG properly represents the system’s components and their interactions. This alignment is crucial for the system’s decisions and actions to be consistent with its overall goals.

\subsubsection{Knowledge Mapping}
In this phase, mechanisms are developed to transform raw data into a structured format compliant with the entities and relationships defined in the concept design stage. Given that the methodology is designed to be applied to any ROS 2 system, it is essential to customize these mechanisms to fit the specific system in use. 

%This process is integral to the representation step. During the knowledge mapping step, the mechanisms needed to transform raw data into a structured format that aligns with the knowledge graph are defined. Since this is aimed to work with any given ROS 2 system, it is also necessary to tailor this mechanisms for the specific ROS 2 system in use

% \subsubsection{Knowledge Insertion}
% A service is included to integrate the knowledge, formatted as nodes and edges, into the knowledge graph. This service ensures that the graph remains an accurate and up-to-date representation of the mission's information.

\subsection{Mission Design}
With the KG structure established, a high-level language is used to specify the mission during this step, including detailed control sequences and queries to interact with the KG. This enables the robot to execute the mission autonomously, with continuous mapping of inputs and outputs to the KG to adapt to changes and make informed decisions. Additionally, during this step, the designer should validate that the specified mission produces the outcomes defined in the definition step.

\section{ROS 2 KG Implementation}

One of the most relevant parts of the methodology is the representation step, which involves the knowledge representation within the KG. To deal with the issues that arise from the handling of knowledge, we have developed different ROS 2 modules that can be used to integrate knowledge graphs into an existing ROS 2 based system. These modules fulfill three main functions:

\begin{enumerate}
    \item \textbf{Knowledge Base:} A ROS 2 node that is in charge of storing the KG data. This node allows for inserting, querying, and deleting both nodes in the KG and edges between them. Additionally, these KG nodes can also store numerical values in terms of \textit{properties}, which can be quite useful in the robotics domain.
    \item \textbf{Knowledge Extractors:} These components are different ROS 2 nodes in charge of interfacing with the current robotic system to be able to extract the relevant knowledge to be added to the KG. Those ROS 2 nodes subscribe to the different available topics and process the published data to generate knowledge. Additionally, they also handle data already contained in the KG to generate new knowledge.
    \item \textbf{Knowledge Retrievers:} These components allow to query the KG about the entities contained within it and the relationships between them.
\end{enumerate}

In terms of software architecture, the knowledge base ROS 2 node centralizes all the information related to the system, while the extractor and retriever ROS 2 nodes interact with it in an N-to-1 fashion during the execution of a given mission. 

Additionally, to handle multi-agent missions, the software generates a local KG for each agent and then merges them into a single KG ensuring there are no duplicate entities. This strategy allows each drone to perform independent missions, reducing read and reaction times. For example, when two drones share airspace, the shared KG benefits mission execution by including entities such as the operator's position or flight status. Furthermore, the drones can infer new knowledge, such as the relative position between them (e.g., "close")

\section{USE CASE: SEARCH AND RESCUE SCENARIO}

In this section, the objective is to test the proposed methodology and the software tools developed during this work in a multi-drone search and rescue mission.

The original robotic system that we will improve using KGs is Aerostack2. Aerostack2 is an open-source software framework designed to create autonomous multi-robot aerial systems. Its modular architecture and multi-robot orientation make it a versatile platform-independent environment capable of addressing a wide range of capabilities for autonomous operation. ROS 2, on the other hand, is an evolution of the popular Robot Operating System (ROS), designed to overcome the limitations of its predecessor. It provides tools, libraries, and conventions for building complex robotic systems and supports multiple programming languages, making it accessible to a wide variety of developers. 

\begin{figure}[htb]
    \centering
    \includegraphics[width=.7\linewidth]{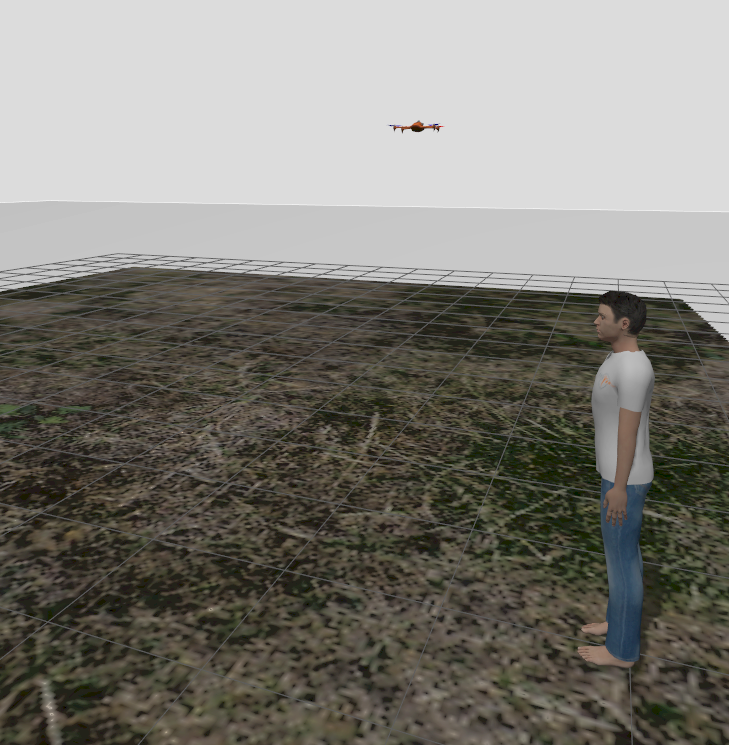}
    \caption{The Gazebo simulation environment.}
    \vspace{-0.3cm}
    \label{fig:Gazebo}
\end{figure}

\begin{figure}[htb]
    \centering
    \includegraphics[width=.95\linewidth]{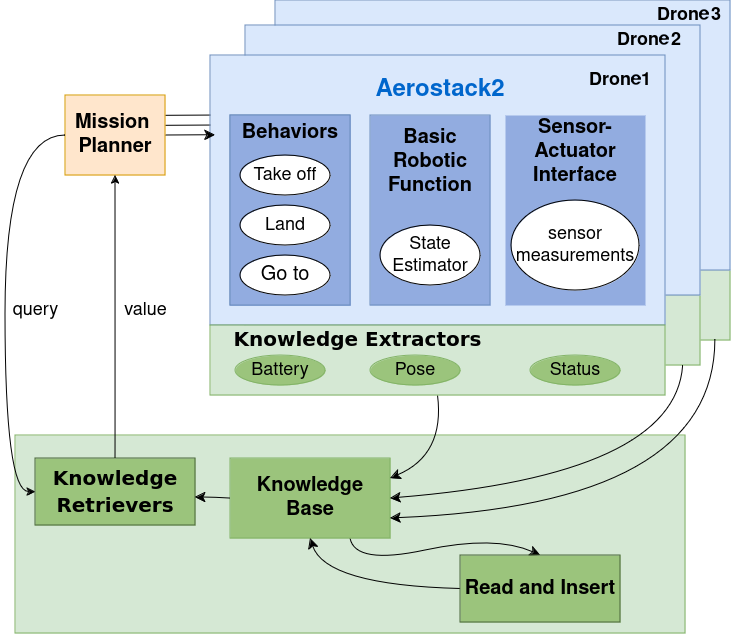}
    \caption{A graphical view of the application of the methodology to enhance Aerostack2. Green components are the ones introduced to the system through the application of the methodology, while blue ones are related to Aerostack2. Knowledge extraction is allocated in each one of the agents, while the knowledge base and the knowledge retrievers are centralized.}
    \label{fig:AS2Integration}
\end{figure}

The presented use case involves a mission where a set of drones must locate a target in an environment. This mission will be simulated in a Gazebo environment as shown in Figure \ref{fig:Gazebo}, with the drones autonomously executing the mission using Aerostack2. Knowledge extractors specifically tailored for Aerostack2 will be employed as described in Section III and Figure \ref{fig:AS2Integration} to enable efficient knowledge handling, integrating the KG capabilities into Aerostack2. This setup will demonstrate how the drones, powered by the advanced knowledge representation and decision-making processes, can effectively carry out the mission in a simulated scenario. 

%The methodology has been developed using ROS 2 to leverage its advanced capabilities and its focus on scalability and interoperability.

\textbf{Definition Stage}: The mission objective is to inspect a specific area and determine the location of a desired object. The requirements are two drones that can execute autonomous flights, able to perceive their environment, and locate the object of interest. Additionally, the drones must have the capability to continuously monitor the state of their battery and their own location.

This mission aims to evaluate the behavior of the KG both when a single agent knowledge is introduced and when the knowledge expected by multiple agents is integrated into a unified graph. This evaluation will allow us to determine the efficiency and effectiveness of the KG in situations with varying levels of complexity and coordination among multiple autonomous agents.

\textbf{Structuring Stage}: The mission can be divided into two main tasks: traversing a defined area and searching for an object, in addition to the tasks responsible for the continuous monitoring of the drones. The full list of identified tasks and sub-tasks is presented in Table \ref{tab:tasks_subtasks}.

\begin{table}[h]
    \renewcommand{\arraystretch}{1.25}
    \centering
    \begin{tabular}[c]{|m{1.5cm}|m{2cm}|m{1.8cm}|m{1.8cm}|}
    \hline
    \textbf{Task} & \textbf{Sub-task} & \textbf{Input} & \textbf{Output} \\
    \hline
    \multirow{2}{1.5cm}{Traverse a defined area} & Determine the current position of the agent & & Current position of each drone \\
    \cline{2-4}
     & Determine the required route to cover the remaining area & Current position of each drone & Remaining path to cover the area \\
    \hline
    \multirow{2}{1.5cm}{Search and localization of the object} & Capture environmental information & & Use onboard cameras \\
    \cline{2-4}
     & Recognize the desired object in the camera image & Camera image & Label the image as \textit{contains} or \textit{does not contain} the object \\
    \hline
    \multirow{3}{1.5cm}{Drone supervision} & Battery status & & High or low level \\
    \cline{2-4}
     & Relative position between drones & \textit{Close} or \textit{not close} & Maintain position or move \\
    \cline{2-4}
     & Navigation status & & Landed/Flying \\
    \hline
    \end{tabular}
    \caption{Tasks and sub-tasks identified for the mission consisting of locating a person in an environment.}
    \label{tab:tasks_subtasks}
    \vspace{-0.15cm}
\end{table}

\textbf{Planning Stage}: Initially, each drone operates independently. The first task is to check its battery status to ensure mission continuity. This check must be performed periodically throughout the mission; if a low battery level is detected, an emergency landing must be carried out if the drone has already taken off, or the takeoff must be prevented if it has not. Additionally, the other drone will take on the responsibility of inspecting the entire area.

After verifying the battery, the drone will take off and head to its inspection area, where it will start sweeping the zone. If it detects the individual, it will send a signal and wait for the other drone to approach so that they can send the individual's coordinates to the operator. Conversely, if the drone does not locate the individual but receives a signal from the other drone, it will halt its trajectory and proceed to the indicated position.

Finally, if neither drone locates the individual, both will complete the inspection of their respective areas and return to the origin station, where they will proceed to land.

\begin{figure}[h]
    \centering
    \includegraphics[width=1\linewidth]{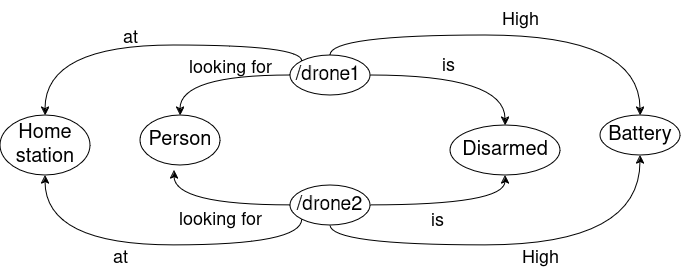}
    \caption{Initial state of the KG. It represents both drones in their respective home stations, with their batteries highly charged and ready to begin the search and rescue mission.}
    \label{fig:initial_graph}
\end{figure}

\textbf{Representation Stage}: 
\begin{enumerate}

    \item \textbf{Knowledge Extraction}: With the necessary parameters and information for successful mission execution determined by an expert technician, the next step is to extract and maintain this knowledge. Utilizing Aerostack2, we can subscribe to relevant topics like position and battery status, ensuring that these data are continuously updated and readily available for accurate and reliable mission execution.

    \item \textbf{Concept Design}: The identified entities in the KG are listed in Table \ref{tab:entities}. Since the relevant knowledge to be stored is related to the current state of each drone, the most important edges between are the ones connecting the drone to the person to indicate if it has located the person, those linking a drone to a status entity to describe its current activity, and the edges linking drones to other drones to indicate proximity and avoid collisions. Table \ref{tab:relationships} outlines all the possible relationships between entities.
    
\begin{table}[h]
    \centering
    \begin{tabular}{|c|c|}
    \hline
    \textbf{Entity} & \textbf{Properties}\\
    \hline
    Drone  & Current pose\\
    \hline
    Battery & Voltage\\
    \hline
    Person & Location\\
    \hline
    Status & Disarmed/Flying/Landed\\
    \hline
    Home station & Location\\
    \hline
    \end{tabular}
    \caption{Entities and properties identified for a search and rescue mission.}
    \vspace{-0.3cm}
    \label{tab:entities}
\end{table}

\begin{table}[h]
    \centering
    \begin{tabular}{|c|c|c|}
    \hline
    \textbf{Source entity}&\textbf{Possible Relationships}&\textbf{Target entity}\\
    \hline
    \multirow{2}*{Drone} & looking for & \multirow{2}*{Person}\\
    &located&\\
    \hline
    Drone & is & Status\\
    \hline
    \multirow{2}*{Drone} & at & \multirow{2}*{Home Station}\\
            &outside&\\
    \hline
    \multirow{3}*{Drone} &High& \multirow{3}*{Battery}\\
     &Medium& \\
    &Low&\\
    \hline
    Drone & close & Drone\\
    \hline
    \end{tabular}
    \caption{Relationships between the different entities identified in a search and rescue mission.}
    \vspace{-0.15cm}
    \label{tab:relationships}
\end{table}

    \item \textbf{Knowledge Mapping}: Once the representation of each part of the information has been defined, specific methods are used to transform the extracted knowledge into a format compatible with the KG. This involves converting the information into entities and relationships that the graph can support. After completing this step, the KG state in the initial situation defined in the mission is shown in Figure \ref{fig:initial_graph}.

\end{enumerate} 

\begin{figure}[h]
    \centering
    \includegraphics[width=1.0\linewidth]{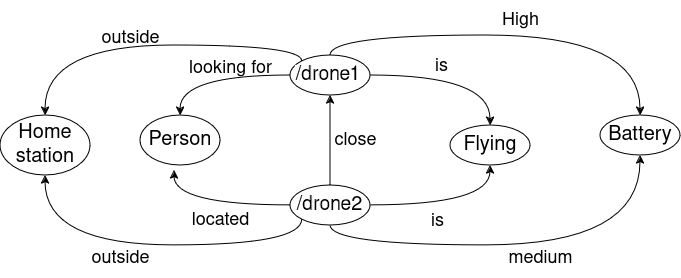}
    \caption{After locating the person, the KG should represent the fact that the two drones are close to each other, that one of them has located the person, and that both of them are still flying.}
    \label{fig:final_graph}
\end{figure}

\textbf{Mission Design Stage}: The planned mission is translated into Python using the tools provided by Aerostack2 and leveraging the capabilities of the KG to perform queries and monitor the current status of the mission at all times.
A rule-based system is used, which analyzes sensitive data through queries and generates consequences. Table \ref{tab:queries} outlines some examples of queries during the mission. Regarding the validation, the final state of the KG after a successful mission is shown in \ref{fig:final_graph}, where the person is located and both drones are close to each other.

    \begin{table}[h]
    \centering
    \begin{tabular}{|m{0.3\linewidth}|c|m{0.3\linewidth}|}
    \hline
    \textbf{Query}&\textbf{Value}&\textbf{Consequences}\\
    \hline
    Battery level & low & Drone return home station\\
 
    \hline
     Inspection status& person located & Send position to home station\\
    \hline
    Relative position between drones & close & Drone moves some distance away\\
    \hline
    \end{tabular}
    \caption{Relevant queries for the search and rescue mission, their return values, and the actions to take in case those values are returned. }
    \label{tab:queries}
    \end{table}
 This demonstrates the advantage of using the knowledge graph, as describing the mission only requires verifying data encoded in a semantic language, rather than acquiring and interpreting numerical data. Figure \ref{fig:final_graph} below shows the knowledge graph when the person has been located.
 
\section{CONCLUSIONS}

The proposed methodology for implementing knowledge graphs in ROS 2 systems offers a robust framework for enhancing knowledge management and decision-making in autonomous missions. By systematically defining, structuring, planning, and representing mission-critical tasks, and by tailoring data extraction methods to specific systems, this approach ensures accurate and efficient integration of knowledge graphs. This integration enables more sophisticated data handling and analysis, ultimately improving the system's ability to make informed decisions autonomously. Through this methodology, robotic systems can achieve greater reliability, adaptability, and performance in complex mission scenarios.

One of the primary challenges of the proposed methodology is that it places the responsibility on the user to manually design how perceived information is mapped to nodes and edges in the knowledge graph. This task requires careful consideration of how system data, such as sensor readings or mission status, should be represented in the graph structure. While this approach provides flexibility, it can also be complex and time-consuming, as it requires a deep understanding of both the system and the knowledge graph to ensure accurate and meaningful representation.

The practical application of this methodology is demonstrated through a mission to locate a person using drones, implemented within the Aerostack2 framework. By employing the proposed steps, from defining mission objectives to mapping data onto a knowledge graph, the system was able to effectively coordinate drone operations and enhance situational awareness. The structured representation of tasks and the tailored data extraction facilitated precise control and real-time decision-making. This implementation underscores the methodology’s potential to improve the operational capabilities of autonomous systems, showcasing its effectiveness in a real-world scenario and highlighting its versatility in handling complex missions within the Aerostack2 framework.

Future work could focus on enhancing the methodology and software components by developing new tools to automate the defined steps, thereby streamlining the entire process. Additionally, incorporating support for reasoning methods beyond rule-based approaches, such as probabilistic or machine learning-based reasoning, could improve decision-making capabilities. Experimenting with different knowledge graph implementations would also be valuable to identify the most efficient solutions for real-time computation, further enhancing the system's performance and responsiveness.

\color{black}
\section*{ACKNOWLEDGMENT}

This work has been supported by the European Union’s Horizon Europe Project No. 101070254 CORESENSE. This work has also been supported by the project INSERTION ref. ID2021-127648OBC32, which was funded by the Spanish Ministry of Science and Innovation under the program "Projects for Knowledge Generating". In addition, this work has also been supported by the project RATEC ref: PDC2022-133643-C22 funded by the Spanish Ministry of Science and Innovation.

%%%%%%%%%%%%%%%%%%%%%%%%%%%%%%%%%%%%%%%%%%%%%%%%%%%%%%%%%%%%%%%%%%%%%%%%%%%%%%%%
% \textcolor{blue}{REVISAR FORMATO BIBLIOGRAFIA}
\bibliographystyle{unsrt}
\bibliography{bibliography}
% \input{bibliography}
% \cite*
%%%%%%%%%%%%%%%%%%%%%%%%%%%%%%%%%%%%%%%%%%%%%%%%%%%%%%%%%%%%%%%%%%%%%%%%%%%%%%%%
%\input{style_guide}
\end{document}